\documentclass{article}

\usepackage[preprint]{neurips_2025}

\usepackage[utf8]{inputenc}
\usepackage[T1]{fontenc}
\usepackage{textcomp} 
\usepackage[dvipsnames]{xcolor}
\usepackage{listings}
\usepackage{hyperref}
\usepackage{url}
\usepackage{booktabs}
\usepackage{amsfonts}
\usepackage{nicefrac}
\usepackage{microtype}
\usepackage{graphicx}

\newcommand{\ProcessThreeDashes}{\textcolor{red}{---}}

\lstdefinelanguage{yaml}{
  keywords={true,false,null,y,n},
  keywordstyle=\color{darkgray}\bfseries,
  basicstyle=\ttfamily\small,
  sensitive=false,
  comment=[l]{\#},
  morecomment=[s]{/*}{*/},
  commentstyle=\color{purple}\ttfamily,
  stringstyle=\color{blue}\ttfamily,
  moredelim=[l][\color{orange}]{\&},
  moredelim=[l][\color{magenta}]{*},
  moredelim=**[il][\color{red}]{:},
  morestring=[b]',
  morestring=[b]",
  literate = {---}{{\ProcessThreeDashes}}3
             {>}{{\textcolor{red}{\textgreater}}}{1}
             {|}{{\textcolor{red}{\textbar}}}{1}
             {\ -\ }{{\mdseries\ -\ }}3,
}

\lstdefinelanguage{json}{
  basicstyle=\normalfont\ttfamily,
  numbers=left,
  numberstyle=\scriptsize,
  stepnumber=1,
  numbersep=8pt,
  showstringspaces=false,
  breaklines=true,
  frame=lines,
  backgroundcolor=\color{gray!10},
  literate=
   *{0}{{{\color{blue}0}}}{1}
    {1}{{{\color{blue}1}}}{1}
    {2}{{{\color{blue}2}}}{1}
    {3}{{{\color{blue}3}}}{1}
    {4}{{{\color{blue}4}}}{1}
    {5}{{{\color{blue}5}}}{1}
    {6}{{{\color{blue}6}}}{1}
    {7}{{{\color{blue}7}}}{1}
    {8}{{{\color{blue}8}}}{1}
    {9}{{{\color{blue}9}}}{1}
    {:}{{{\color{red}{:}}}}{1}
    {,}{{{\color{red}{,}}}}{1}
    {\{}{{{\color{red}{\{}}}}{1}
    {\}}{{{\color{red}{\}}}}}{1}
    {[}{{{\color{red}{[}}}}{1}
    {]}{{{\color{red}{]}}}}{1},
}

\title{CORTEX: Collaborative LLM Agents for High-Stakes Alert Triage}

\author{
  Bowen Wei \\
  George Mason University \\
  \texttt{bwei2@gmu.edu}
  \And
  Yuan Shen Tay \\
  Fluency Security \\
  \texttt{yuanshen.tay@fluencysecurity.com}
  \And
  Howard Liu \\
  Fluency Security \\
  \texttt{howard@fluencysecurity.com}
  \And
  Jinhao Pan \\
  George Mason University \\
  \texttt{jpan23@gmu.edu}
  \And
  Kun Luo \\
  Fluency Security \\
  \texttt{kun@fluencysecurity.com}
  \And
  Ziwei Zhu \\
  George Mason University \\
  \texttt{zzhu20@gmu.edu}
  \And
  Chris Jordan \\
  Fluency Security \\
  \texttt{chris@fluencysecurity.com}
}

\begin{document}
\maketitle

\begin{abstract}
Security Operations Centers (SOCs) are overwhelmed by tens of thousands of daily alerts, of which only a small fraction correspond to genuine attacks. This overload creates alert fatigue, leading to overlooked threats and analyst burnout. Classical detection pipelines are brittle and context-poor, while recent LLM-based approaches typically rely on a single model to interpret logs, retrieve context, and adjudicate alerts end-to-end—an approach that struggles with noisy enterprise data and offers limited transparency. We propose \emph{CORTEX}, a multi-agent LLM architecture for high-stakes alert triage in which specialized agents collaborate over real evidence: a behavior-analysis agent inspects activity sequences, evidence-gathering agents query external systems, and a reasoning agent synthesizes findings into an auditable decision. To support training and evaluation, we release a dataset of fine-grained SOC investigations from production environments, capturing step-by-step analyst actions and linked tool outputs. Across diverse enterprise scenarios, CORTEX substantially reduces false positives and improves investigation quality over state-of-the-art single-agent LLMs.
\end{abstract}

\section{Introduction}
Security Operations Centers (SOCs) form the first line of defense against enterprise attacks, yet they are overwhelmed by an onslaught of alerts—often tens of thousands per day—of which only a small fraction indicate genuine threats. Empirical studies report false-positive rates approaching 99\%~\cite{alahmadi2022false}, creating extreme alert fatigue: critical signals are easily overlooked (as in the Target breach~\cite{riley2014missed,finkle2014target}), while analysts face burnout~\cite{tines2023voice}. Traditional pipelines based on rules and anomaly detectors are brittle and context-poor, flooding operators with noise rather than insight. Recent industry reports estimate that 40--45\% of enterprise alerts are false positives~\cite{orca2022cloud,esg2023fastly}, underscoring the urgent need for more precise, transparent triage.

Large language models (LLMs) have been explored for summarizing incidents and assisting analysts~\cite{zhang2025cybersurvey,deng2024pentestgpt}, but most approaches adopt a single-agent paradigm: one model must interpret logs, retrieve context, and adjudicate alerts end-to-end. Even with chain-of-thought prompting~\cite{wei2022chain} or ReAct-style tool use~\cite{yao2022react}, such models often falter on long-horizon, high-stakes investigations and provide limited auditability. This gap is particularly problematic in security-critical domains, where decisions must be both accurate and explainable~\cite{arrieta2020explainable}.

We address this challenge with a \emph{divide-and-conquer multi-agent architecture} for SOC triage. Specialized agents assume distinct roles: a \emph{Behavior Analysis Agent} identifies relevant investigative workflows; \emph{Evidence Acquisition Agents} ground hypotheses by querying external systems (e.g., SIEM logs, threat intelligence); and a \emph{Reasoning \& Coordination Agent} synthesizes evidence into a transparent triage decision. Agents communicate through structured messages, iteratively cross-checking claims, akin to human analyst teams~\cite{chen2023agentverse,du2023improving,liang2024encouraging}.

To support training and evaluation, we release a dataset of \emph{fine-grained SOC workflows}, collected from production environments across more than ten scenarios. Unlike prior datasets that provide only coarse labels, ours captures full investigative traces—stepwise analyst actions, tool queries and outputs, and intermediate observations—enabling both training and evaluation of process-level reasoning.

Experiments across diverse enterprise scenarios show that CORTEX substantially reduces false positives and improves investigative quality compared to single-agent LLMs.

\paragraph{Contributions.}
\begin{itemize}
    \item \textbf{CORTEX:} a role-specialized, tool-using, auditable multi-agent architecture for SOC triage.
    \item \textbf{Fine-grained SOC workflow dataset:} process-level triage traces across 10+ real scenarios.
    \item \textbf{Empirical validation:} large reductions in false positives and improved reasoning quality over baselines.
\end{itemize}

\section{Related Work}
\subsection{LLMs in Cybersecurity}
LLMs are increasingly applied across defensive and offensive cybersecurity. Surveys synthesize hundreds of works spanning threat-intelligence extraction, knowledge-graph reasoning, vulnerability analysis, and attack automation \cite{zhang2025cybersurvey}. On defense, systems organize attack knowledge (e.g., \textsc{AttacKG+}) and build TI knowledge graphs from unstructured text \cite{zhang2025attackgplus,hu2024llmtikg}. On offense, \textsc{PentestGPT} demonstrates automated penetration testing with tool use and iterative planning, evaluated on real systems and CTFs \cite{deng2024pentestgpt}. The OWASP Top 10 for LLM Applications formalizes risks (prompt injection, data poisoning, model DoS) relevant to both red and blue teams \cite{owasp2024top10}. Field reports further document SOC burnout and workflow pain points that motivate automation \cite{tines2023voice}. Compared to these threads, far fewer works model end-to-end SOC alert investigation as a \emph{tool-grounded, role-specialized} reasoning pipeline, and fewer still explore \emph{distillation} of that process into a deployable single model. Our work targets precisely this gap.

\subsection{Alert Fatigue and Automated Triage in SOCs}
SOCs contend with extreme alert volume and high false-positive rates, imposing substantial manual validation~\cite{alahmadi2022false,orca2022cloud}. The Target breach illustrates the operational risk of drowning in noise despite vendor alerts~\cite{riley2014missed,finkle2014target}. Long breach lifecycles further motivate faster, higher-precision triage (241 days to identify \emph{and} contain, per the 2025 IBM report)~\cite{ibm2025breach}. Classical automation frames triage as supervised prioritization or learning-to-rank over prior analyst decisions. Representative systems report meaningful workload reductions with low false-negative rates: \textsc{AACT} imitates analyst actions to auto-close benign alerts and escalate critical ones~\cite{turcotte2025aact}; \textsc{AlertPro} leverages contextual features and reinforcement learning to rank alerts and improve multi-step scenario handling~\cite{gong2024alertpro}. Yet, purely statistical models can be brittle to novel patterns and demand substantial, continuously curated histories. In parallel, emerging “agentic” SOC platforms (e.g., Radiant Security; Dropzone AI) advertise end-to-end investigation coverage and large speedups~\cite{radiant2025platform,radiant2025press,dropzone2024blog,dropzone2025guide}, but typically incur notable computational costs. These trends together motivate SOC-specific triage that mirrors analyst roles, \emph{grounds} decisions in tool-fetched evidence, and remains efficient at deployment scale—precisely the design goals of our multi-agent, tool-using approach.

\subsection{Collaborative Multi-Agent LLM Systems}
Multiple LLM agents—with specialized roles, communication, and verification—can outperform single models on complex reasoning. Multi-agent debate iteratively critiques candidate answers to improve factuality and consistency~\cite{du2023improving,liang2024encouraging}. General frameworks such as \textsc{AgentVerse}~\cite{chen2023agentverse}, \textsc{CAMEL}~\cite{li2023camel}, \textsc{MetaGPT}~\cite{hong2023metagpt}, and \textsc{AutoGen}~\cite{wu2023autogen} provide orchestration patterns (roles, turn-taking, tool calls) that enable task decomposition and cross-checking. Empirical analyses also catalog failure modes—specification errors, inter-agent misalignment, and weak termination/verification—highlighting the need for principled protocols and reliability checks~\cite{cemri2025failure}. This literature motivates domain-aligned roles, disciplined message passing, and verification anchored in external evidence, while also surfacing a practical requirement: retain collaborative benefits without inflating inference cost or coordination complexity. Our architecture instantiates SOC-aligned roles and structured evidence exchange.

\section{Methods}
Our approach integrates (i) a divide-and-conquer multi-agent architecture for triage and (ii) a fine-grained SOC workflow dataset for process-level supervision.

\subsection{CORTEX Architecture}
\begin{figure}[t]
\centering
\includegraphics[width=0.9\linewidth]{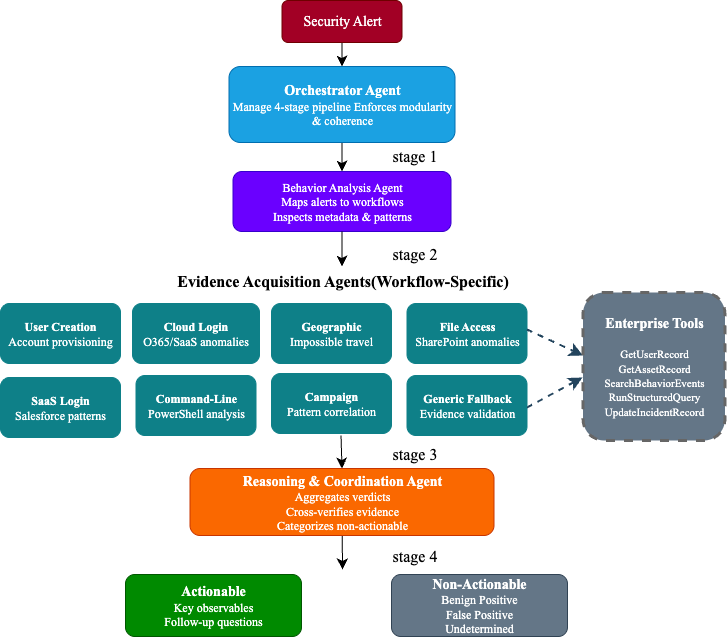}
\caption{\textbf{CORTEX architecture.} A security alert enters a four-stage pipeline. \emph{Stage 1: Orchestrator Agent} manages execution and modularity. \emph{Stage 2: Behavior Analysis Agent} maps alerts to workflows. \emph{Stage 3: Evidence Acquisition Agents} (workflow-specific) query enterprise tools (e.g., SIEM, identity, asset context) using typed APIs to validate hypotheses. \emph{Stage 4: Reasoning \& Coordination Agent} aggregates workflow outputs, cross-verifies evidence, applies conservative escalation logic, and emits a structured, auditable report with observables and follow-ups.}
\label{fig:cortex_arch}
\end{figure}

\noindent\textbf{Roles.} 
CORTEX decomposes triage into four stages. The \emph{Orchestrator Agent} manages the pipeline, enforces modularity, and ensures coherent handoffs between roles. The \emph{Behavior Analysis Agent} routes alerts to the most relevant workflows. Workflow-specific \emph{Evidence Acquisition Agents} execute calibrated playbooks by querying enterprise tools. The \emph{Reasoning \& Coordination Agent} reconciles evidence and produces a structured, auditable report (see Fig.~\ref{fig:cortex_arch}).

\noindent\textbf{Pipeline.} 
The pipeline unfolds in four stages: (1) \emph{orchestration} (execution control and consistency checks); (2) \emph{classification} (routing) by the Behavior Analysis Agent; (3) \emph{workflow analysis} with calibrated, evidence-grounded criteria executed by Evidence Acquisition Agents; and (4) \emph{synthesis \& actionability extraction} in the Reasoning \& Coordination Agent, which finalizes the verdict, extracts observables, and proposes follow-ups. We instantiate seven workflows from production SOC data—credential changes, anomalous logins, file access anomalies, geographic impossibility, SaaS irregularities, user creation, and command-line execution (Table~\ref{tab:thresholds}).

\begin{table}[t]
\centering
\caption{Workflow-specific actionability criteria (examples).}
\label{tab:thresholds}
\begin{tabular}{lll}
\toprule
Workflow & Decision Logic & Actionable Condition \\
\midrule
Credential Change & Auth modifications & Removal/addition by established user \\
User Creation & Account provisioning & Elevated privileges granted \\
Cloud Login Anomaly & Risk + history & Score $>1000$ with recent activity \\
Geo Impossibility & Distance/time & $>$500 miles apart, infeasible time \\
File Access Anomaly & Risk + location & Score $>1000$ from uncommon site \\
SaaS Login Anomaly & Rule triggers & $\geq 3$ abnormal triggers \\
Command-Line Exec. & Command analysis & Malicious code by admin user \\
\bottomrule
\end{tabular}
\end{table}

\noindent\textbf{Reporting.} 
CORTEX adopts a conservative decision policy: if any workflow escalates, the overall verdict is \emph{actionable}. Otherwise, alerts are assigned to interpretable non-actionable categories (Table~\ref{tab:nonactionable}). Reports for actionable alerts extract observables (e.g., IPs, accounts) and recommend follow-up questions.

\begin{table}[t]
\centering
\caption{Non-actionable categories.}
\label{tab:nonactionable}
\begin{tabular}{ll}
\toprule
Category & Definition \\
\midrule
Benign Positive & Suspicious but expected behavior \\
False Positive—Logic & Due to flawed detection rule \\
False Positive—Data & Due to data errors \\
Undetermined & Insufficient evidence \\
\bottomrule
\end{tabular}
\end{table}

\noindent\textbf{Tooling.} 
Typed tools ground reasoning (Table~\ref{tab:tools}); examples include \texttt{getUserRecord}, \texttt{searchBehaviorEvents}, and \texttt{runStructuredQuery}. These abstractions standardize access to logs, user and asset records, and parametric queries, ensuring decisions remain auditable and reproducible.

\begin{table}[t]
\centering
\caption{Representative tools.}
\label{tab:tools}
\begin{tabular}{lll}
\toprule
Tool & Description & Example Workflow \\
\midrule
\texttt{getUserRecord} & Retrieve user attributes & Credential Change, User Creation \\
\texttt{getAssetRecord} & Endpoint context & Command-Line Execution \\
\texttt{searchBehaviorEvents} & Raw logs & Generic \\
\texttt{searchBehaviorSummaries} & Aggregates & Geo, Cloud Login \\
\texttt{runStructuredQuery} & Parametric reports & Cloud, File, SaaS Login \\
\texttt{updateIncidentRecord} & Write status & Reporting \\
\bottomrule
\end{tabular}
\end{table}

\subsection{Fine-Grained SOC Workflow Dataset}
\label{sec:dataset}

\noindent\textbf{Construction Protocol.}  
We collected end-to-end investigations from enterprise SOCs across ten scenarios, spanning cloud identity, SaaS file access, endpoint detections, and IAM policy changes. For each alert, the dataset records raw telemetry, analyst actions, tool queries, intermediate reasoning, and final adjudication. Sensitive fields (e.g., usernames, hostnames, IPs) are pseudonymized while preserving structural integrity. All traces are serialized as JSON with a consistent schema, enabling both supervised training and structured evaluation.

\noindent\textbf{Schema.}  
Each JSON trace contains a unique \texttt{id}; the investigated \texttt{entity} (e.g., user account, IAM role, or endpoint); metadata such as \texttt{account}, \texttt{tenant}, \texttt{timestamp} (Unix epoch), \texttt{time\_iso} (ISO 8601); and an overall \texttt{riskScore}. The core of each record is the \texttt{properties} object, which holds one or more triggered behavioral rules. A rule specifies the \texttt{behaviorRule} name, a human-readable \texttt{description}, an \texttt{attributes} dictionary of evidence fields, a local \texttt{riskScore}, and optional \texttt{risks} tags. Attributes cover identity (\texttt{Username}, \texttt{ARN}, \texttt{UserType}); network context (\texttt{ClientIP}, \texttt{ActorIP}, \texttt{City}, \texttt{Country}, \texttt{ISP}); system context (\texttt{OS}, \texttt{BrowserType}, \texttt{Hostname}, \texttt{Workload}); operational semantics (\texttt{Operation}, \texttt{EventName}, \texttt{CmdLine}, \texttt{ParentProcess}); artifacts (\texttt{FileName}, \texttt{ExploitPath}); and security annotations (\texttt{MFA}, \texttt{Severity}, \texttt{Remediation}, \texttt{Verdict}).

\noindent\textbf{Examples.}  
An O365 login trace records IP geolocation, ISP, OS, and first-seen login flags. An AWS IAM modification trace captures IAM roles and API calls (e.g., \texttt{PutRolePolicy}) with associated user-agent strings. A Defender uncommon-activity trace highlights guest account additions in Azure AD groups, annotated with severity and verdict. Endpoint threat-control traces log PowerShell command lines, parent processes, and remediation outcomes. These heterogeneous sources are normalized into a unified schema while retaining domain-specific detail.

\noindent\textbf{Features and Labels.}  
From each record we derive contextual features (user, asset, environment identifiers), behavioral features (operations, commands, file-access patterns), and security annotations (analytic risk scores, anomaly tags). Each investigation is labeled with a triage outcome: \emph{Actionable} (escalated to incident) or \emph{Non-actionable}, further subclassed into \emph{Benign Positive}, \emph{False Positive—Logic}, \emph{False Positive—Data}, or \emph{Undetermined}. This supports both coarse- and fine-grained evaluation.

\noindent\textbf{Scale.}  
The dataset comprises several thousand traces. Each trace typically contains two to four behavioral rules and six to twelve attributes. Coverage spans cloud (Azure AD, AWS), SaaS (OneDrive, SharePoint), and endpoint sources, reflecting the multi-signal nature of real SOC alerts.

\section{Experiments}

\subsection{Experimental Setup}

\noindent\textbf{Datasets.}
All experiments use the fine-grained SOC workflow dataset described in Section~\ref{sec:dataset}. All PII is pseudonymized upstream as part of the dataset.

\noindent\textbf{Tasks and Outputs.}
Given an alert trace (JSON), systems must produce a schema-valid triage report with: (i) a binary verdict (Actionable / Non-actionable), (ii) a non-actionable subclass when applicable (Benign Positive, False Positive—Logic, False Positive—Data, Undetermined), (iii) a brief rationale grounded in fetched evidence, and (iv) extracted observables (e.g., user, IP, file, asset). Outputs must comply with a fixed JSON schema (Appendix, Listing~A.1).

\noindent\textbf{Baselines (Single-Agent).}
We compare two single-model settings:
\emph{Prompt-only.} A single LLM consumes the alert JSON and emits the triage report without tool calls. Prompts include task instructions, schema constraints, and few-shot exemplars per workflow.
\emph{ReAct-style tool use.} A single LLM plans and executes tool calls (same typed tools as CORTEX) via a ReAct prompt. Tool budget and turn caps are matched to CORTEX for fairness. The model must ground claims in returned tool outputs and emit a schema-valid report.

\noindent\textbf{CORTEX Configuration (Ours).}
CORTEX instantiates the Orchestrator, Behavior Analysis, workflow-specific Evidence Acquisition Agents, and the Reasoning \& Coordination Agent (Fig.~\ref{fig:cortex_arch}). The Behavior Analysis Agent routes to one or more workflows; Evidence Agents execute calibrated checks (Table~\ref{tab:thresholds}); the Reasoning Agent applies conservative “escalate-on-any” synthesis and composes the structured report. All agents share the same tool library used by the ReAct baseline.

\noindent\textbf{Implementation.}
We implement all agents using the \emph{OpenAI Agents SDK} under the \emph{Model Context Protocol (MCP)}. Each typed tool (e.g., \texttt{getUserRecord}, \texttt{searchBehaviorEvents}, \texttt{runStructuredQuery}) is exposed as an MCP \emph{tool} with JSON-schema arguments and deterministic JSON returns; logs and auxiliary artifacts are exposed as MCP \emph{resources}. Inter-agent communication occurs via MCP sessions with per-alert, ephemeral context; cross-alert state is disabled to prevent leakage. The ReAct baseline is implemented over the same SDK and MCP tool adapters to ensure parity; measured latencies therefore include SDK/MCP overhead uniformly across systems. All MCP traces are logged for auditability.

\noindent\textbf{Evaluation Metrics.}
\emph{Decision quality:} macro-F1 over Actionable/Non-actionable; subclass macro-F1 over the four non-actionable categories; false-positive rate (FPR) computed on non-actionable predictions that disagree with ground truth; recall computed on actionable alerts. \emph{Efficiency:} output tokens, tool calls, and end-to-end latency .

\subsection{Triage Performance}
Table~\ref{tab:triage_performance} reports classification quality across models. 
CORTEX achieves the strongest overall performance, improving actionable F1 by $+0.12$ over the best single-agent baseline (0.66 $\rightarrow$ 0.78) and reducing the false-positive rate by $10.7$ points (24.9\% $\rightarrow$ 14.2\%). 
Macro-F1 across actionable/non-actionable decisions reaches $0.82$, and subclass F1 increases by $+0.15$, reflecting sharper distinctions among benign positives, logic-driven false positives, and data-driven false positives.

\begin{table}[t]
\centering
\caption{Decision performance on the test set. FPR is computed on non-actionable predictions.}
\label{tab:triage_performance}
\begin{tabular}{lcccc}
\toprule
Model & Act. F1 & Non-act. F1 & Subclass F1 & FPR (\%) \\
\midrule
Single-agent (prompt)   & 0.61 & 0.73 & 0.49 & 29.8 \\
Single-agent (tool-use) & 0.66 & 0.77 & 0.54 & 24.9 \\
\textbf{CORTEX (ours)}  & \textbf{0.78} & \textbf{0.86} & \textbf{0.69} & \textbf{14.2} \\
\bottomrule
\end{tabular}
\end{table}

\subsection{Efficiency}
CORTEX trades additional coordination for higher decision quality while remaining within our target SOC triage SLO of $\sim$3\,min per full ticket. Its median end-to-end time is $152.4$\,s ($\approx 2.54$\,min), compared to $44.6$\,s for the single-agent ReAct-style baseline and $28$\,s for prompt-only. This corresponds to a $+107.8$\,s increase over the tool-using baseline ($+241.7\%$, $3.42\times$ slower) and $+124.4$\,s over prompt-only ($+444.3\%$, $5.44\times$ slower). The primary driver is the larger token footprint introduced by multi-agent message passing and the serialization of richer tool outputs: CORTEX processes $23{,}600$ tokens vs.\ $4{,}152$ for the tool-using baseline ($+468.4\%$, $5.68\times$). Average tool calls rise only modestly (from $1.3$ to $3.1$, $\Delta{=}1.8$, $2.38\times$), indicating that the latency gap is not solely due to more API hits but also to increased deliberation and inter-agent exchange. These efficiency costs accompany the accuracy gains reported in Table~\ref{tab:triage_performance} (higher F1, lower FPR) and yield more auditable, evidence-grounded investigations.

\begin{table}[t]
\centering
\caption{Efficiency comparison. Latency is median end-to-end time per alert (full ticket resolution).}
\label{tab:efficiency}
\begin{tabular}{lccc}
\toprule
Model & Tokens & Tool Calls & Latency (s) \\
\midrule
Single-agent (prompt)   & 2{,}100 & 0.0 & 28 \\
Single-agent (tool-use) & 4{,}152 & 1.3 & 44.6 \\
\textbf{CORTEX (ours)}  & \textbf{23{,}600} & \textbf{3.1} & \textbf{152.4} \\
\bottomrule
\end{tabular}
\end{table}

\section{Conclusion}
We introduced \textsc{CORTEX}, a collaborative, tool-grounded multi-agent architecture for high-stakes SOC alert triage. By decomposing the task across role-specialized agents and constraining each step to operate over typed tools and auditable artifacts, \textsc{CORTEX} improves both decision quality and transparency relative to single-agent baselines. On our evaluation suite, \textsc{CORTEX} increases actionable F1 from 0.66 to 0.78 and reduces false positives from 24.9\% to 14.2\% while maintaining operationally acceptable end-to-end latency. Beyond outcome metrics, \textsc{CORTEX} produces structured reports with explicit evidence links, enabling downstream review, compliance, and post-incident learning.

A second contribution is a fine-grained SOC workflow dataset that captures full investigative traces—alerts, tool queries, intermediate observations, and final adjudications—across diverse enterprise scenarios. This process-level supervision supports training agents to follow disciplined playbooks rather than relying solely on outcome labels, and it enables new measurements of reasoning fidelity (step accuracy, tool-policy match, and grounding consistency).

\paragraph{Limitations and Future Work.}
Our evaluation is limited by dataset coverage (ten-plus scenarios) and the availability and quality of upstream telemetry. Like other agentic systems, \textsc{CORTEX} can be sensitive to distribution shift, prompt injection, or incomplete context returned by tools. Future directions include (i) stronger termination and verification protocols (e.g., cross-checking with learned critics), (ii) adaptive tool budgeting and scheduling across agents, (iii) distillation of multi-agent traces into a compact single-model policy for cost/latency reduction, (iv) continual learning from analyst feedback and A/B tests, and (v) expanded benchmarks for red-team robustness and privacy-preserving operation.

Overall, \textsc{CORTEX} offers a practical template for auditable, role-specialized LLM agents in security operations. We hope the architecture, evaluation protocol, and released dataset catalyze further research on reliable, efficient agents for safety-critical domains.

\bibliography{refs}

\begin{thebibliography}{28}
\providecommand{\natexlab}[1]{#1}
\providecommand{\url}[1]{\texttt{#1}}
\expandafter\ifx\csname urlstyle\endcsname\relax
  \providecommand{\doi}[1]{doi: #1}\else
  \providecommand{\doi}{doi: \begingroup \urlstyle{rm}\Url}\fi

\bibitem[AlAhmadi et~al.(2022)AlAhmadi, Axon, and Martinovic]{alahmadi2022false}
Bushra~A AlAhmadi, Louise Axon, and Ivan Martinovic.
\newblock 99\% false positives: A qualitative study of {SOC} analysts' perspectives on security alarms.
\newblock \emph{arXiv preprint arXiv:2210.01649}, 2022.

\bibitem[Arrieta et~al.(2020)Arrieta, D{\'\i}az-Rodr{\'\i}guez, Del~Ser, Bennetot, Tabik, Barbado, Garc{\'\i}a, Gil-L{\'o}pez, Molina, Benjamins, et~al.]{arrieta2020explainable}
Alejandro~Barredo Arrieta, Natalia D{\'\i}az-Rodr{\'\i}guez, Javier Del~Ser, Adrien Bennetot, Siham Tabik, Alberto Barbado, Salvador Garc{\'\i}a, Sergio Gil-L{\'o}pez, Daniel Molina, Richard Benjamins, et~al.
\newblock Explainable artificial intelligence (xai): Concepts, taxonomies, opportunities and challenges toward responsible ai.
\newblock \emph{Information fusion}, 58:\penalty0 82--115, 2020.

\bibitem[Cemri et~al.(2025)Cemri, Zhang, Smith, et~al.]{cemri2025failure}
Emre Cemri, Yanchen Zhang, Kevin Smith, et~al.
\newblock Why do multi-agent {LLM} systems fail?
\newblock In \emph{Conference Proceedings}, 2025.

\bibitem[Chen et~al.(2024)Chen, Su, Zuo, Yang, Yuan, Chan, Yu, Lu, Hung, Qian, et~al.]{chen2023agentverse}
Weize Chen, Yusheng Su, Jingwei Zuo, Cheng Yang, Chenfei Yuan, Chi-Min Chan, Heyang Yu, Yaxi Lu, Yi-Hsin Hung, Chen Qian, et~al.
\newblock {AgentVerse}: Facilitating multi-agent collaboration and exploring emergent behaviors.
\newblock In \emph{International Conference on Learning Representations}, 2024.

\bibitem[Deng et~al.(2024)Deng, Liu, Mayoral-Vilches, Liu, Li, Xu, Zhang, Liu, Pinzger, and Rass]{deng2024pentestgpt}
Gelei Deng, Yi~Liu, V{\'\i}ctor Mayoral-Vilches, Peng Liu, Yuekang Li, Yuan Xu, Tianwei Zhang, Yang Liu, Martin Pinzger, and Stefan Rass.
\newblock {PentestGPT}: Evaluating and harnessing large language models for automated penetration testing.
\newblock In \emph{Proceedings of the 33rd USENIX Security Symposium}, 2024.

\bibitem[{Dropzone AI}(2024)]{dropzone2024blog}
{Dropzone AI}.
\newblock The rise of {AI SOC} analysts: Transforming security operations.
\newblock \url{https://www.dropzone.ai/blog/}, 2024.

\bibitem[{Dropzone AI}(2025)]{dropzone2025guide}
{Dropzone AI}.
\newblock Alert triage in 2025: The complete guide to 90\% faster investigations, 2025.

\bibitem[Du et~al.(2023)Du, Li, Torralba, Tenenbaum, and Mordatch]{du2023improving}
Yilun Du, Shuang Li, Antonio Torralba, Joshua~B. Tenenbaum, and Igor Mordatch.
\newblock Improving factuality and reasoning in language models through multiagent debate.
\newblock In \emph{International Conference on Machine Learning}, 2023.

\bibitem[{Enterprise Strategy Group}(2023)]{esg2023fastly}
{Enterprise Strategy Group}.
\newblock Web application and {API} security survey.
\newblock Technical report, ESG for Fastly, 2023.

\bibitem[Finkle and Heavey(2014)]{finkle2014target}
Jim Finkle and Susan Heavey.
\newblock Target says it declined to act on early alert of cyber breach.
\newblock \emph{Reuters}, 2014.

\bibitem[Gong et~al.(2024)Gong, Wang, Yang, and Liang]{gong2024alertpro}
Xiaorui Gong, Xiaoyu Wang, Xiaobo Yang, and Xueping Liang.
\newblock Combating alert fatigue with {AlertPro}: Context-aware alert prioritization using reinforcement learning for multi-step attack detection.
\newblock \emph{Computers \& Security}, 137:\penalty0 103621, 2024.

\bibitem[Hong et~al.(2023)Hong, Zhuge, Chen, Xiong, Ming, Zha, Zhang, Wen, Zhang, Zheng, et~al.]{hong2023metagpt}
Sirui Hong, Mingchen Zhuge, Jonathan Chen, Xiawu Xiong, Yuheng Ming, Ceyao Zha, Jinlin Zhang, Chenglin Wen, Yiqi Zhang, Kunlun Zheng, et~al.
\newblock {MetaGPT}: Meta programming for a multi-agent collaborative framework.
\newblock \emph{arXiv preprint arXiv:2308.00352}, 2023.

\bibitem[Hu et~al.(2024)Hu, Zhang, Li, and Chen]{hu2024llmtikg}
Xiaojun Hu, Yang Zhang, Jing Li, and Haixin Chen.
\newblock Building threat intelligence knowledge graphs with large language models.
\newblock \emph{IEEE Transactions on Information Forensics and Security}, 19:\penalty0 2345--2358, 2024.

\bibitem[{IBM Security}(2025)]{ibm2025breach}
{IBM Security}.
\newblock Cost of a data breach report 2025.
\newblock Technical report, IBM Corporation, 2025.

\bibitem[Li et~al.(2023)Li, Hammoud, Itani, Khizbullin, and Ghanem]{li2023camel}
Guohao Li, Hasan Abed Al~Kader Hammoud, Hani Itani, Dmitrii Khizbullin, and Bernard Ghanem.
\newblock {CAMEL}: Communicative agents for ``mind'' exploration of large language model society.
\newblock In \emph{Advances in Neural Information Processing Systems}, 2023.

\bibitem[Liang et~al.(2024)Liang, He, Jiao, Wang, Wang, Wang, Yang, Shi, and Tu]{liang2024encouraging}
Tian Liang, Zhiwei He, Wenxiang Jiao, Xing Wang, Yan Wang, Rui Wang, Yujiu Yang, Shuming Shi, and Zhaopeng Tu.
\newblock Encouraging divergent thinking in large language models through multi-agent debate.
\newblock In \emph{Proceedings of the 2024 Conference on Empirical Methods in Natural Language Processing}, pages 17889--17904, 2024.

\bibitem[{Orca Security}(2022)]{orca2022cloud}
{Orca Security}.
\newblock 2022 cloud security alert fatigue report.
\newblock Technical report, Orca Security, 2022.

\bibitem[{OWASP Foundation}(2024)]{owasp2024top10}
{OWASP Foundation}.
\newblock {OWASP} top 10 for large language model applications.
\newblock \url{https://owasp.org/www-project-top-10-for-large-language-model-applications/}, 2024.

\bibitem[{Radiant Security}(2025{\natexlab{a}})]{radiant2025platform}
{Radiant Security}.
\newblock Radiant security: {AI}-powered {SOC} automation platform.
\newblock \url{https://radiantsecurity.ai/}, 2025{\natexlab{a}}.

\bibitem[{Radiant Security}(2025{\natexlab{b}})]{radiant2025press}
{Radiant Security}.
\newblock Radiant security announces breakthrough in autonomous {SOC} operations.
\newblock Press Release, 2025{\natexlab{b}}.

\bibitem[Riley et~al.(2014)Riley, Elgin, Lawrence, and Matlack]{riley2014missed}
Michael Riley, Ben Elgin, Dune Lawrence, and Carol Matlack.
\newblock Missed alarms and 40 million stolen credit card numbers: How {Target} blew it.
\newblock \emph{Bloomberg Businessweek}, 13:\penalty0 1--7, 2014.

\bibitem[{Tines}(2023)]{tines2023voice}
{Tines}.
\newblock Voice of the {SOC} report.
\newblock Technical report, Tines, 2023.

\bibitem[Turcotte et~al.(2025)Turcotte, Labr{\`e}che, and Paquette]{turcotte2025aact}
Melissa J.~M. Turcotte, Fran{\c{c}}ois Labr{\`e}che, and Serge-Olivier Paquette.
\newblock Automated alert classification and triage ({AACT}): An intelligent system for the prioritisation of cybersecurity alerts.
\newblock \emph{arXiv preprint arXiv:2505.09843}, 2025.

\bibitem[Wei et~al.(2022)Wei, Wang, Schuurmans, Bosma, Xia, Chi, Le, Zhou, et~al.]{wei2022chain}
Jason Wei, Xuezhi Wang, Dale Schuurmans, Maarten Bosma, Fei Xia, Ed~Chi, Quoc~V Le, Denny Zhou, et~al.
\newblock Chain-of-thought prompting elicits reasoning in large language models.
\newblock In \emph{Advances in neural information processing systems}, volume~35, pages 24824--24837, 2022.

\bibitem[Wu et~al.(2023)Wu, Bansal, Zhang, Wu, Li, Zhu, Jiang, Zhang, Zhang, Liu, et~al.]{wu2023autogen}
Qingyun Wu, Gagan Bansal, Jieyu Zhang, Yiran Wu, Beibin Li, Erkang Zhu, Li~Jiang, Xiaoyun Zhang, Shaokun Zhang, Jiale Liu, et~al.
\newblock {AutoGen}: Enabling next-gen {LLM} applications via multi-agent conversation.
\newblock \emph{arXiv preprint arXiv:2308.08155}, 2023.

\bibitem[Yao et~al.(2023)Yao, Zhao, Yu, Du, Shafran, Narasimhan, and Cao]{yao2022react}
Shunyu Yao, Jeffrey Zhao, Dian Yu, Nan Du, Izhak Shafran, Karthik Narasimhan, and Yuan Cao.
\newblock React: Synergizing reasoning and acting in language models.
\newblock In \emph{International Conference on Learning Representations}, 2023.

\bibitem[Zhang et~al.(2025{\natexlab{a}})Zhang, Bu, Wen, Liu, Fei, Xi, Li, Yang, Zhu, and Meng]{zhang2025cybersurvey}
Jie Zhang, Haoyu Bu, Hui Wen, Yongji Liu, Haiqiang Fei, Rongrong Xi, Lun Li, Yun Yang, Hongsong Zhu, and Dan Meng.
\newblock When {LLMs} meet cybersecurity: A systematic literature review.
\newblock \emph{Cybersecurity}, 8\penalty0 (55), 2025{\natexlab{a}}.

\bibitem[Zhang et~al.(2025{\natexlab{b}})Zhang, Liu, Chen, and Wang]{zhang2025attackgplus}
Qian Zhang, Wei Liu, Ming Chen, and Xiaoyu Wang.
\newblock {AttacKG+}: Enhancing attack knowledge graphs with large language models.
\newblock In \emph{IEEE Symposium on Security and Privacy}, 2025{\natexlab{b}}.

\end{thebibliography}
\bibliographystyle{plainnat}

\appendix

\section{Workflow Prompts}
\label{app:prompts}

This appendix provides the full YAML definitions for all workflow prompts used in \textsc{CORTEX}. 
To align with the main paper, we use the \emph{agent} roles \textbf{Orchestrator}, \textbf{Behavior Analysis Agent}, \textbf{Evidence Acquisition Agents}, and \textbf{Reasoning \& Coordination Agent}; and the \emph{typed tools} \texttt{getUserRecord}, \texttt{getAssetRecord}, \texttt{searchBehaviorEvents}, \texttt{searchBehaviorSummaries}, \texttt{runStructuredQuery}, \texttt{updateIncidentRecord}. 
Decision policies (e.g., escalation thresholds) are applied by the \textbf{Reasoning \& Coordination Agent} rather than via a separate ``Action\_Detection\_Agent'' tool.

\begin{table}[h]
\small
\centering
\caption{Summary of workflow prompts in \textsc{CORTEX}. Decision policies are enforced by the Reasoning \& Coordination Agent; evidence is fetched via typed tools.}
\label{tab:workflow_summary}
\begin{tabular}{lll}
\toprule
\textbf{Workflow} & \textbf{Detection Goal} & \textbf{Key Evidence / Tools} \\
\midrule
Add User & Suspicious user creation/update & \texttt{getUserRecord} (roles), policy thresholds \\
Authentication Change & Auth method changes & \texttt{getUserRecord}, policy (remove/add/change) \\
Coro & Coro vendor signals & behaviorRule extraction, policy=escalate \\
Generic & Fallback analysis & \texttt{searchBehaviorEvents}, policy triage \\
Multiple ISP & Impossible travel & \texttt{runStructuredQuery}(`GetRecentLoginActivity`) \\
O365 Guest & Guest account activity & \texttt{getUserRecord} (guest roles) \\
O365 Login & Risky login & \texttt{runStructuredQuery}(`GetRecentHighRiskActivity`) \\
PowerShell & Endpoint execution & \texttt{searchBehaviorEvents}, \texttt{getAssetRecord} \\
Salesforce Abnormal Login & Risky Salesforce login & \texttt{runStructuredQuery}(`GetRecentRuleActivity`) \\
SharePoint File & Risky file access & risk score from event \\
\bottomrule
\end{tabular}
\end{table}

\subsection{Add User Workflow}
\lstset{language=Yaml, basicstyle=\ttfamily\footnotesize, breaklines=true, columns=fullflexible}
\begin{lstlisting}
name: Add_User_Workflow
description: >
  Analyze events that create or update users. Match when behavior rules contain:
  "User_Added", "User_Updated", or "Add_Member_To_Group".
model: gpt-5-mini
prompt: |
  Role: Orchestrator for an O365 user add/update event (JSON).

  Steps:
  1) Extract target user email(s) from 'properties.*.TargetUser'. For each target user:
     - Evidence Acquisition Agent invokes typed tool `getUserRecord(email, account, tenant)`.
     - If no email or record, mark target_user_record = "Unknown".
  2) Determine 'target_user_admin' from returned user record(s) (roles include admin-like privileges).
     - This classification is performed by the Reasoning & Coordination Agent (no external tool).
  3) Reasoning & Coordination Agent applies policy:
     - If any target_user_admin = Admin => actionable = true; else actionable = false.
  4) Emit schema-valid JSON as below.

  Output JSON schema:
  {
    "report": {
      "target_user_record": "<Found/Unknown>",
      "target_user_admin": "<Admin/User/Unknown>",
      "reasoning_target_user_admin": "<string>"
    },
    "actionable": <bool>,
    "reasoning": "<string>",
    "summary": "<string>"
  }

tools:
  - name: getUserRecord
    description: Retrieve a user's directory record for roles/attributes.
\end{lstlisting}

\subsection{Authentication Change Workflow}
\begin{lstlisting}
name: Authentication_Change_Workflow
description: >
  Analyze authentication method changes: add, remove, or modify MFA/password.
  Match when behavior rules contain: "Add_Authentication_Method" or "Remove_Authentication_Method".
model: gpt-5-mini

prompt: |
  Role: Orchestrator for an authentication method change event (JSON).

  Steps:
  1) Identity: user email = input.entity; (account, tenant) from input.
  2) Evidence Acquisition Agent calls `getUserRecord(email, account, tenant)`.
     - If not found: report.user_record = "Unknown"; new_user = "Unknown"; goto Step 4.
  3) If found: Reasoning & Coordination Agent computes account age vs time_iso to set:
     new_user = Yes/No/Unknown, with justification.
  4) Policy (Reasoning & Coordination Agent):
     - Removing method => actionable = true.
     - Not removing, and new_user = Yes => actionable = false.
     - Not removing, and new_user = No or Unknown => actionable = true.
  5) Emit schema-valid JSON.

  Output JSON schema:
  {
    "report": {
      "user_record": "<Found/Unknown>",
      "new_user": "<Yes/No/Unknown>",
      "reasoning_new_user": "<string>"
    },
    "actionable": <bool>,
    "reasoning": "<string>",
    "summary": "<string>"
  }

tools:
  - name: getUserRecord
    description: Retrieve a user's directory record for roles/attributes.
\end{lstlisting}

\subsection{Correlation (Coro) Workflow}
\begin{lstlisting}
name: Coro_Workflow
description: >
  Match Coro vendor events ("Coro_*") and escalate per policy.
model: gpt-5-mini

prompt: |
  Role: Orchestrator for a Coro event (JSON).

  Steps:
  1) Extract user_email = input.entity; list behavior_rule names from input.
  2) Policy: Reasoning & Coordination Agent marks actionable = true for vendor Coro events.
  3) Emit schema-valid JSON with behavior_rules included.

  Output JSON schema:
  {
    "report": { "user_email": "<string>", "behavior_rules": ["<string>", ...] },
    "actionable": true,
    "reasoning": "Escalated per Coro vendor policy.",
    "summary": "<string>"
  }
\end{lstlisting}

\subsection{Generic Workflow}
\begin{lstlisting}
name: Generic_Workflow
description: >
  Fallback workflow for alerts that do not match any specific workflow.
model: gpt-5-mini

prompt: |
  Role: Orchestrator coordinating Evidence Acquisition and Reasoning.

  Steps:
  1) Evidence Acquisition Agent may call `searchBehaviorEvents(...)` for relevant raw logs,
     if and only if the input lacks sufficient direct evidence fields.
  2) Reasoning & Coordination Agent validates event evidence and determines actionability:
     - CLOSE_TICKET (invalid/not actionable)
     - ESCALATE_TO_TIER_TWO (valid/actionable)
     - REQUIRES_ADDITIONAL_INFO (insufficient/conflicting)
  3) Emit schema-valid JSON.

  Output JSON schema:
  {
    "report":{
      "validation": true/false,
      "validation_reasoning": "<string>",
      "recommendation": "CLOSE_TICKET" | "ESCALATE_TO_TIER_TWO" | "REQUIRES_ADDITIONAL_INFO"
    },
    "actionable": true/false,
    "reasoning": "<string>",
    "summary": "<string>"
  }

tools:
  - name: searchBehaviorEvents
    description: Query raw telemetry for supporting evidence.
\end{lstlisting}

\subsection{Multiple ISP (Impossible Travel) Workflow}
\begin{lstlisting}
name: MultipleISP_Workflow
description: >
  Determine whether observed logins constitute impossible travel.
  Match when behavior rule includes "Multiple_ISPs" or multiple ISPs under an O365 login rule.
model: gpt-5-mini

prompt: |
  Role: Orchestrator for impossible-travel evaluation.

  Steps:
  1) Evidence Acquisition Agent calls `runStructuredQuery(GetRecentLoginActivity, account+tenant, key=entity, time_iso)`
     to obtain recent login activity with geo/ISP fields.
  2) Reasoning & Coordination Agent evaluates distance/time between logins (8-hour window),
     considering ISP diversity and plausible explanations (VPN/mobile). 
     Sets report.impossible_travel (bool) with detailed reasoning.
  3) Policy: actionable = report.impossible_travel.
  4) Emit JSON.

  Output JSON schema:
  {
    "report": {
      "impossible_travel": <bool>,
      "impossible_travel_reasoning": "<string>"
    },
    "actionable": <bool>,
    "reasoning": "<string>",
    "summary": "<string>"
  }

tools:
  - name: runStructuredQuery
    description: Execute a parametric report (e.g., GetRecentLoginActivity) and return JSON rows.
\end{lstlisting}

\subsection{Office 365 Guest Workflow}
\begin{lstlisting}
name: Office365_Guest_Workflow
description: >
  Analyze guest user activity (key formatted as <username>#ext#@<tenant>.onmicrosoft.com).
model: gpt-5-mini

prompt: |
  Role: Orchestrator for guest user activity.

  Steps:
  1) Evidence Acquisition Agent calls `getUserRecord(guest_email, account, tenant)`.
     - If not found: guest_user_record = "Unknown"; guest_user_admin = "Unknown".
  2) Reasoning & Coordination Agent inspects roles in record (if found) to classify:
     guest_user_admin = Admin/User.
  3) Policy: actionable = (guest_user_admin == Admin).
  4) Emit JSON.

  Output JSON schema:
  {
    "report": {
      "guest_user_record": "<Found/Unknown>",
      "guest_user_admin": "<Admin/User/Unknown>",
      "reasoning_guest_user_admin": "<string>"
    },
    "actionable": <bool>,
    "reasoning": "<string>",
    "summary": "<string>"
  }

tools:
  - name: getUserRecord
    description: Retrieve guest user directory record for role evaluation.
\end{lstlisting}

\subsection{Office 365 Login Workflow}
\begin{lstlisting}
name: Office365_Login_Workflow
description: >
  Analyze O365 login rule with risk and recent high-risk context.
model: gpt-5-mini

prompt: |
  Role: Orchestrator for O365 login event.

  Steps:
  1) user_email = input.entity.
  2) Evidence Acquisition Agent calls
     `runStructuredQuery(GetRecentHighRiskActivity, account+tenant, key=user_email, time_iso)`
     to count high-risk activities (rowCount) and keep 'row' JSON.
  3) Extract the risk score for the O365 Login behavior rule (not the ticket total).
  4) Reasoning & Coordination Agent applies policy:
     - risk <= 1000 => actionable = false
     - risk > 1000 & recent_high_risk_count == 0 => actionable = false
     - risk > 1000 & recent_high_risk_count > 0 => actionable = true
  5) Emit JSON.

  Output JSON schema:
  {
    "report": {
      "user_email": "<string>",
      "recent_activity_riskScore_greater_than_2000_count": <int>,
      "high_risk_activity_raw_json_row": "<string>"
    },
    "actionable": <bool>,
    "reasoning": "<string>",
    "summary": "<string>"
  }

tools:
  - name: runStructuredQuery
    description: Execute `GetRecentHighRiskActivity` and return rows and rowCount.
\end{lstlisting}

\subsection{PowerShell Workflow}
\begin{lstlisting}
name: Powershell_Workflow
description: >
  Analyze PowerShell execution for malicious behavior and remediation status.
model: gpt-5-mini

prompt: |
  Role: Orchestrator for PowerShell event.

  Steps:
  1) Evidence Acquisition Agent reads event 'attributeSummaries' (no external tool),
     classifies code as Malicious / Non-Malicious with one-sentence rationale.
  2) Evidence Acquisition Agent checks disinfection status from event fields
     (e.g., status/actionTaken indicates 'disinfected' => Disinfect; else Non-Disinfect).
  3) Reasoning & Coordination Agent applies policy:
     actionable = (powerShell_Malicious == true AND user_has_admin == true).
     (Admin status may be derived via `getUserRecord` if user context provided.)
  4) Emit JSON.

  Output JSON schema:
  {
    "report": {
      "powerShell_Malicious": <bool>,
      "reasoning": "<string>",
      "dis_Infect_Detection": "<Disinfect/Non-Disinfect>",
      "reasoning_Dis_Infect": "<string>"
    },
    "actionable": <bool>,
    "reasoning": "<string>",
    "summary": "<string>"
  }

tools:
  - name: getUserRecord
    description: Retrieve user record if admin status is required for policy.
\end{lstlisting}

\subsection{Salesforce Abnormal Login Workflow}
\begin{lstlisting}
name: Salesforce_Abnormal_Login_Workflow
description: >
  Analyze Salesforce abnormal login rule with recent rules context.
model: gpt-5-mini

prompt: |
  Role: Orchestrator for Salesforce abnormal login event.

  Steps:
  1) user_email = input.entity.
  2) Evidence Acquisition Agent calls
     `runStructuredQuery(GetRecentRuleActivity, account+tenant, rule="Fluency_Salesforce_Login_Status_Abnormal", key=user_email, time_iso)`
     and obtains recent_rule_count = rowCount.
  3) Reasoning & Coordination Agent policy:
     - recent_rule_count < 3 => actionable = false
     - recent_rule_count >= 3 => actionable = true
  4) Emit JSON.

  Output JSON schema:
  {
    "report": {
      "user_email": "<string>",
      "recent_rule_count": <int>
    },
    "actionable": <bool>,
    "reasoning": "<string>",
    "summary": "<string>"
  }

tools:
  - name: runStructuredQuery
    description: Execute `GetRecentRuleActivity` with the named rule and return rowCount.
\end{lstlisting}

\subsection{SharePoint File Workflow}
\begin{lstlisting}
name: Sharepoint_File_Workflow
description: >
  Analyze SharePoint file access/download risk based on rule risk score.
model: gpt-5-mini

prompt: |
  Role: Orchestrator for SharePoint file event.

  Steps:
  1) Extract the risk score for the SharePoint File behavior rule from input.properties.
  2) Reasoning & Coordination Agent policy:
     - risk <= 1000 => actionable = false
     - risk > 1000  => actionable = true
  3) Emit JSON.

  Output JSON schema:
  {
    "report": {
      "sharepoint_risk_score": <int>
    },
    "actionable": <bool>,
    "reasoning": "<string>",
    "summary": "<string>"
  }

tools: []
\end{lstlisting}

\section{Case Studies}
\label{app:case-studies}

We present five case studies illustrating how \textsc{CORTEX} processes alerts. Each includes the agent reasoning trace and the final JSON report.

\subsection{Case Study 1: Add User Workflow}
\textbf{Event:} New user added to a tenant with elevated privileges.

\textbf{Reasoning Trace:}
\begin{itemize}
  \item \textbf{Behavior Analysis Agent}: Matches \texttt{User\_Added} $\rightarrow$ routes to Add User Workflow.
  \item \textbf{Evidence Acquisition Agents}: Invoke \texttt{getUserRecord(jane.doe@corp.com)}; record found.
  \item \textbf{Reasoning \& Coordination Agent}: Roles include \texttt{GlobalAdmin} $\rightarrow$ target\_user\_admin=Admin; policy $\Rightarrow$ actionable.
\end{itemize}

\textbf{Final Report:}
\begin{lstlisting}[language=json]
{
  "report": {
    "target_user_record": "Found",
    "target_user_admin": "Admin",
    "reasoning_target_user_admin": "Target user roles include GlobalAdmin in the retrieved user record."
  },
  "actionable": true,
  "reasoning": "Admin privilege assignment is actionable under policy.",
  "summary": "New user provisioned with Global Admin privileges; escalate for immediate review."
}
\end{lstlisting}

\subsection{Case Study 2: Authentication Change Workflow}
\textbf{Event:} A long-standing user removed MFA and switched to password login.

\textbf{Reasoning Trace:}
\begin{itemize}
  \item \textbf{Behavior Analysis Agent}: Matches \texttt{Remove\_Authentication\_Method}.
  \item \textbf{Evidence Acquisition Agents}: \texttt{getUserRecord(john.smith@corp.com)}; createdDateTime=2019.
  \item \textbf{Reasoning \& Coordination Agent}: new\_user=No; removal policy $\Rightarrow$ actionable.
\end{itemize}

\textbf{Final Report:}
\begin{lstlisting}[language=json]
{
  "report": {
    "user_record": "Found",
    "new_user": "No",
    "reasoning_new_user": "User account created in 2019 (older than 30 days)."
  },
  "actionable": true,
  "reasoning": "Removal of an authentication method is always actionable.",
  "summary": "Established user removed MFA; flagged for potential account compromise."
}
\end{lstlisting}

\subsection{Case Study 3: Multiple ISP / Geo-Impossibility}
\textbf{Event:} User logged in from London and New York within 30 minutes.

\textbf{Reasoning Trace:}
\begin{itemize}
  \item \textbf{Behavior Analysis Agent}: Matches Multiple ISPs.
  \item \textbf{Evidence Acquisition Agents}: \texttt{runStructuredQuery(GetRecentLoginActivity,...)} returns London 10:15 UTC (BT UK) and NYC 10:45 UTC (Verizon).
  \item \textbf{Reasoning \& Coordination Agent}: $\sim$5{,}500 km in 30 min $\Rightarrow$ impossible travel $\Rightarrow$ actionable.
\end{itemize}

\textbf{Final Report:}
\begin{lstlisting}[language=json]
{
  "report": {
    "impossible_travel": true,
    "impossible_travel_reasoning": "Logins at 10:15 UTC (London) and 10:45 UTC (New York) are ~5,500 km apart within 30 minutes, exceeding feasible travel limits."
  },
  "actionable": true,
  "reasoning": "Pattern reflects impossible travel within the evaluation window.",
  "summary": "User exhibited impossible travel (London to NYC within 30 minutes); escalate."
}
\end{lstlisting}

\subsection{Case Study 4: Office 365 Guest Workflow}
\textbf{Event:} Guest user invited into Finance group.

\textbf{Reasoning Trace:}
\begin{itemize}
  \item \textbf{Behavior Analysis Agent}: Matches Guest user pattern.
  \item \textbf{Evidence Acquisition Agents}: \texttt{getUserRecord(jane.guest\#ext@othercorp.com)}; record found.
  \item \textbf{Reasoning \& Coordination Agent}: roles=User $\Rightarrow$ not actionable.
\end{itemize}

\textbf{Final Report:}
\begin{lstlisting}[language=json]
{
  "report": {
    "guest_user_record": "Found",
    "guest_user_admin": "User",
    "reasoning_guest_user_admin": "Guest user record shows no admin roles; roles indicate standard user access."
  },
  "actionable": false,
  "reasoning": "Guest does not hold admin privileges under policy.",
  "summary": "Guest added to Finance group with no admin privileges; not actionable."
}
\end{lstlisting}

\subsection{Case Study 5: PowerShell Workflow}
\textbf{Event:} PowerShell execution with encoded command attempting persistence.

\textbf{Reasoning Trace:}
\begin{itemize}
  \item \textbf{Behavior Analysis Agent}: Matches PowerShell execution.
  \item \textbf{Evidence Acquisition Agents}: Inspect \texttt{attributeSummaries} $\Rightarrow$ \texttt{-EncodedCommand} and registry persistence $\Rightarrow$ Malicious; disinfection status = Disinfect.
  \item \textbf{Reasoning \& Coordination Agent}: If user is Admin (via \texttt{getUserRecord}), policy $\Rightarrow$ actionable despite disinfection.
\end{itemize}

\textbf{Final Report:}
\begin{lstlisting}[language=json]
{
  "report": {
    "powerShell_Malicious": true,
    "reasoning": "Encoded command and registry-based persistence consistent with malicious behavior.",
    "dis_Infect_Detection": "Disinfect",
    "reasoning_Dis_Infect": "Endpoint telemetry indicates disinfection completed."
  },
  "actionable": true,
  "reasoning": "Malicious PowerShell executed by an Admin meets escalation policy.",
  "summary": "Admin executed malicious PowerShell with persistence; disinfected but escalated for follow-up."
}
\end{lstlisting}
\end{document}